\documentclass[letterpaper]{article} % DO NOT CHANGE THIS
\usepackage[draft]{aaai2026}  % DO NOT CHANGE THIS
\usepackage{times}  % DO NOT CHANGE THIS
\usepackage{helvet}  % DO NOT CHANGE THIS
\usepackage{courier}  % DO NOT CHANGE THIS
\usepackage[hyphens]{url}  % DO NOT CHANGE THIS
\usepackage{graphicx} % DO NOT CHANGE THIS
\usepackage{natbib}  % DO NOT CHANGE THIS AND DO NOT ADD ANY OPTIONS TO IT
\usepackage{caption} % DO NOT CHANGE THIS AND DO NOT ADD ANY OPTIONS TO IT
\usepackage{algorithm}
\usepackage{algorithmic}

\usepackage{amsmath,amsfonts}
\usepackage{array}
\usepackage{subfig}
\usepackage{textcomp}
\usepackage{url}
\usepackage{verbatim}
\usepackage{graphicx}
\usepackage{cite}
\usepackage{xcolor}
\usepackage{soul}
\usepackage{amssymb}
\usepackage{enumitem}

\usepackage{arydshln}
\usepackage{fancyhdr}
\usepackage{varwidth}
\usepackage{arydshln}
\usepackage{multirow}
\usepackage{mathtools}
\usepackage{algorithmic}
\usepackage{algorithm}
\usepackage{tabularray}
\usepackage{datetime2}

\usepackage{newfloat}
\usepackage{listings}
\usepackage{pifont}

\usepackage{kotex}
\usepackage{soul}

\newcommand{\sys}{TCA}

\newcommand{\remove}[1]{}

\DeclareMathOperator*{\argmax}{arg\,max}

\usepackage{newfloat}
\usepackage{listings}
\DeclareCaptionStyle{ruled}{labelfont=normalfont,labelsep=colon,strut=off} % DO NOT CHANGE THIS
\floatstyle{ruled}
\newfloat{listing}{tb}{lst}{}
\floatname{listing}{Listing}
\title{Timestep-Compressed Attack on Spiking Neural Networks through Timestep-Level Backpropagation}
\author{
    Donghwa Kang\textsuperscript{\rm 1}, Doohyun Kim\textsuperscript{\rm 1}, Sang-Ki Ko\textsuperscript{\rm 2}, Jinkyu Lee\textsuperscript{\rm 3,}\thanks{Jinkyu Lee initiated this work while affiliated with Sungkyunkwan University.}, Hyeongboo Baek\textsuperscript{\rm 2}, \\
    Brent ByungHoon Kang\textsuperscript{\rm 1}
}
\affiliations{
    \textsuperscript{\rm 1}Korea Advanced Institute of Science and Technology (KAIST) \\
    \textsuperscript{\rm 2}University of Seoul \\
    \textsuperscript{\rm 3}Yonsei University
}

\usepackage{bibentry}
\begin{document}

\maketitle
\begin{abstract}

State-of-the-art (SOTA) gradient-based adversarial attacks on spiking neural networks (SNNs), which largely rely on extending FGSM and PGD frameworks, face a critical limitation: substantial attack latency from multi-timestep processing, rendering them infeasible for practical real-time applications. 
This inefficiency stems from their design as direct extensions of ANN paradigms, which fail to exploit key SNN properties. 
In this paper, we propose the timestep-compressed attack (TCA), a novel framework that significantly reduces attack latency. 
TCA introduces two components founded on key insights into SNN behavior. 
First, timestep-level backpropagation (TLBP) is based on our finding that global temporal information in backpropagation to generate perturbations is not critical for an attack's success, enabling per-timestep evaluation for early stopping. 
Second, adversarial membrane potential reuse (A-MPR) is motivated by the observation that initial timesteps are inefficiently spent accumulating membrane potential, a warm-up phase that can be pre-calculated and reused. 
Our experiments on VGG-11 and ResNet-17 with the CIFAR-10/100 and CIFAR10-DVS datasets show that TCA significantly reduces the required attack latency by up to 56.6\% and 57.1\% compared to SOTA methods in white-box and black-box settings, respectively, while maintaining a comparable attack success rate.

\end{abstract}
    
\section{Introduction}
\label{sec:intro}

Spiking neural networks (SNNs) are often regarded as the third generation of artificial neural networks, known for
their event-driven design and strong similarity to biological processes. 
Unlike conventional artificial neural networks (ANNs), SNNs use binary sequences to communicate between spiking neurons, producing spikes only when the membrane potential exceeds a certain threshold during the inference over multiple timesteps. This unique feature makes SNNs well-suited for handling spatio-temporal data and helps reduce power consumption~\cite{TGK19}. 
Beyond image classification, SNNs have shown effectiveness in tasks such as text recognition and object detection.~\cite{SCH23, ZZL23}

Despite their general resilience~\cite{SRP20}, the vulnerability of SNNs to adversarial attacks poses a significant security risk. This has led to the development of gradient-based attacks, which have evolved along two main paths. 
The first involves extending de facto standard attack methods (i.e., FGSM~\cite{GSS14}, PGD~\cite{KGB18}) from ANNs to the SNN domain. 
The second focuses on developing SNN-specific gradient generation methods to accommodate their unique temporal dynamics. Key techniques include STBP~\cite{WDL18}, which adapts backpropagation through time (BPTT) for SNNs; RGA~\cite{BDH23}, which approximates gradients based on firing rates; and HART~\cite{HBS23}, a hybrid approach integrating both temporal and rate information. 
Crucially, these SNN-specific methods primarily serve to compute the gradients that are then fed into the established FGSM and PGD frameworks to craft the final attack.

\begin{figure}
    \centering
    \includegraphics[width=1\linewidth]{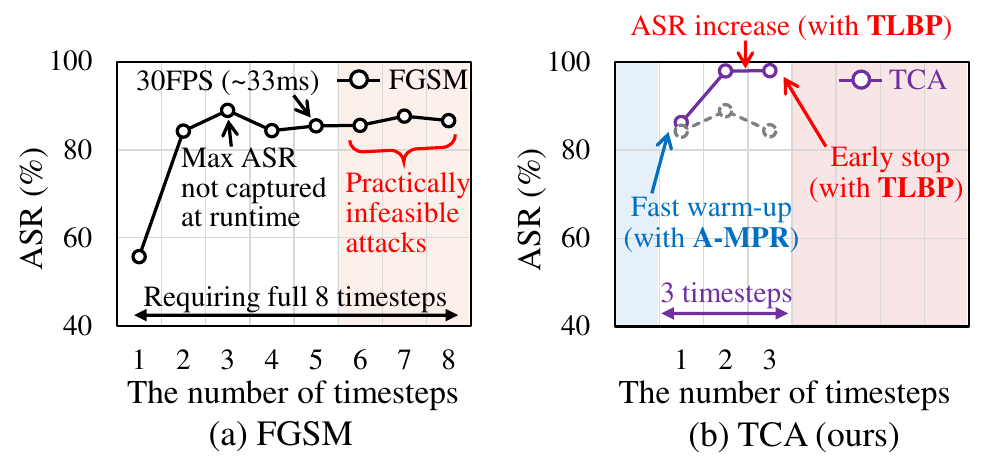}
    \caption{ASR over the number of timesteps for (a) conventional FGSM and (b) TCA, both utilizing STBP, on CIFAR-10 for VGG-11. 
    TCA uses A-MPR for a fast warm-up and TLBP for early stopping to achieve high ASR with fewer timesteps compared to FGSM.}
    \label{fig:intro}
\end{figure}

A critical limitation of SNN attack methods exploiting FGSM or PGD is the substantial latency from multi-timestep processing; for instance, the STBP-based attack in Fig.~\ref{fig:intro}(a) requires eight full forward-backward passes to generate a single adversarial perturbation. 
While the specific ``real-time" deadline depends on the application, a processing rate of 30 FPS is often considered the threshold for real-time performance in many fields, including object recognition in autonomous vehicles or CCTV systems~\cite{ZSD25, SWG24, CKP24}. 
Under such a widely used criterion, the latency of an eight-timestep attack becomes practically infeasible as it exceeds the operational deadline; PGD-based attacks are even slower. 
This suggests that SOTA attacks on SNNs remain largely theoretical, stemming from their design as direct extensions of ANN paradigms that fail to leverage two SNN-specific properties. 
First, there is a distinct latency-ASR (attack success ratio) trade-off (as shown in Fig.~\ref{fig:intro}) which is also highly input-dependent; the number of timesteps needed to succeed varies significantly across inputs. 
This presents the challenge of developing a runtime mechanism to capture this input-dependent vulnerability for early stopping. Second, this trade-off is non-linear, as ASR often increases sharply in the initial timesteps. 
This poses a further challenge of how to selectively allocate computational resources across different attack stages to minimize overall latency.

In this paper, we propose a novel timestep-compressed attack (\sys{}) that establishes such a new paradigm, circumventing the reliance on full-timestep information to significantly reduce the required number of timesteps, illustrated in Fig.~\ref{fig:intro}(b). 
\sys{} incorporates two primary components targeting a classification problem. 
First, timestep-level backpropagation (TLBP) employs a per-timestep strategy for generating perturbations (while eliminating BPTT) to the input image. 
This approach enables cross-entropy monitoring at each timestep for early stopping and, counter-intuitively, significantly improves the final ASR compared to the existing FGSM and PGD using full-timestep temporal information. 
Second, adversarial membrane potential reuse (A-MPR) facilitates a rapid warm-up by initializing the attack with pre-calculated membrane potentials. 
This ``membrane image" is generated offline via a purpose-built loss function that balances gradient accuracy with adversarial efficacy to ensure the extraction of effective gradients from the onset of the attack. 

We implemented TCA to all STBP, RGA, and HART, and it supports both white-box and black-box attacks.
In line with the SOTA studies, we conducted experiments on CIFAR-10/100 and CIFAR10-DVS for VGG-11 and ResNet-17. 
Our experiments validate TCA's effectiveness on VGG-11 and ResNet-17 across the CIFAR-10 and CIFAR-100 datasets. While preserving a comparable ASR, TCA lowers the required attack latency by up to 56.6\% in white-box and 57.1\% in black-box scenarios, respectively, against SOTA methods.

The main contributions of this paper are as follows:
\begin{itemize}
    \item We analyze the impact of global temporal information within the backpropagation process of gradient-based attacks, revealing its minimal effect on the final ASR.
    \item Based on this analysis, we propose TLBP, a novel per-window backpropagation method that enables input-dependent early stopping for low-latency attacks.
    \item We propose A-MPR, an offline, rapid warm-up technique that pre-calculates and reuses adversarial membrane potentials to overcome the inefficiency of early attack timesteps.
    \item We demonstrate the effectiveness of our framework through extensive experiments, showing a superior latency-ASR trade-off compared to SOTA methods in both white-box and black-box scenarios.
\end{itemize}

\section{Related Work}
\label{sec:related_work}

\paragraph{Learning method of SNNs. }
The two primary methods for training SNNs are ANN-SNN conversion~\cite{CCK15} and direct training~\cite{WDL18}. 
ANN-SNN conversion involves training a conventional ANN and then mapping its weights to an SNN of the same architecture. While converted SNNs can achieve comparable accuracy to ANNs without retraining, they typically require a large number of timesteps for inference, though recent works have sought to mitigate this~\cite{DSG21, DNB15}. 
Alternatively, direct training uses methods like STBP to train the SNN from scratch. 
STBP was proposed by Wu et al.~\cite{WDL18} who recognized the similarity between the temporal dynamics of SNNs and RNNs, adapting the BPTT~\cite{WPJ02} algorithm. 
Direct training with STBP enables high accuracy with fewer timesteps than conversion methods, but traditionally struggled to match the peak accuracy of ANNs. 
However, recent studies have made progress in closing this accuracy gap~\cite{FYC21, LDP16, NMZ19}.

\paragraph{Adversarial attacks for SNNs. }
The vulnerability of ANNs to adversarial examples—subtle, imperceptible input perturbations that cause misclassification—was first demonstrated by Szegedy et al.~\cite{SZS13}. 
This led to the development of gradient-based attack methods such as FGSM~\cite{GSS14} and PGD~\cite{KGB18}, which generate perturbations using the model's gradients. 
These methods have since been extended to SNNs, typically using the STBP gradient to guide the perturbation~\cite{SPS19}. However, the non-differentiable nature of spiking neurons makes direct gradient calculation inefficient. 
To overcome this, alternative backpropagation methods have been developed that approximate gradients using firing rates or other temporal information. 
For example, RGA~\cite{BDH23} approximates gradients based on firing ratios, while HART~\cite{HBS23} uses a hybrid of firing ratio and temporal information. 
A common drawback of these approaches, however, is that their reliance on the FGSM and PGD frameworks necessitates performing a forward and/or backward pass for each timestep in the sequence.

\section{Preliminaries}
\label{sec:preliminaries}

\begin{figure}[t!]
    \centering
    \includegraphics[width=1\linewidth]{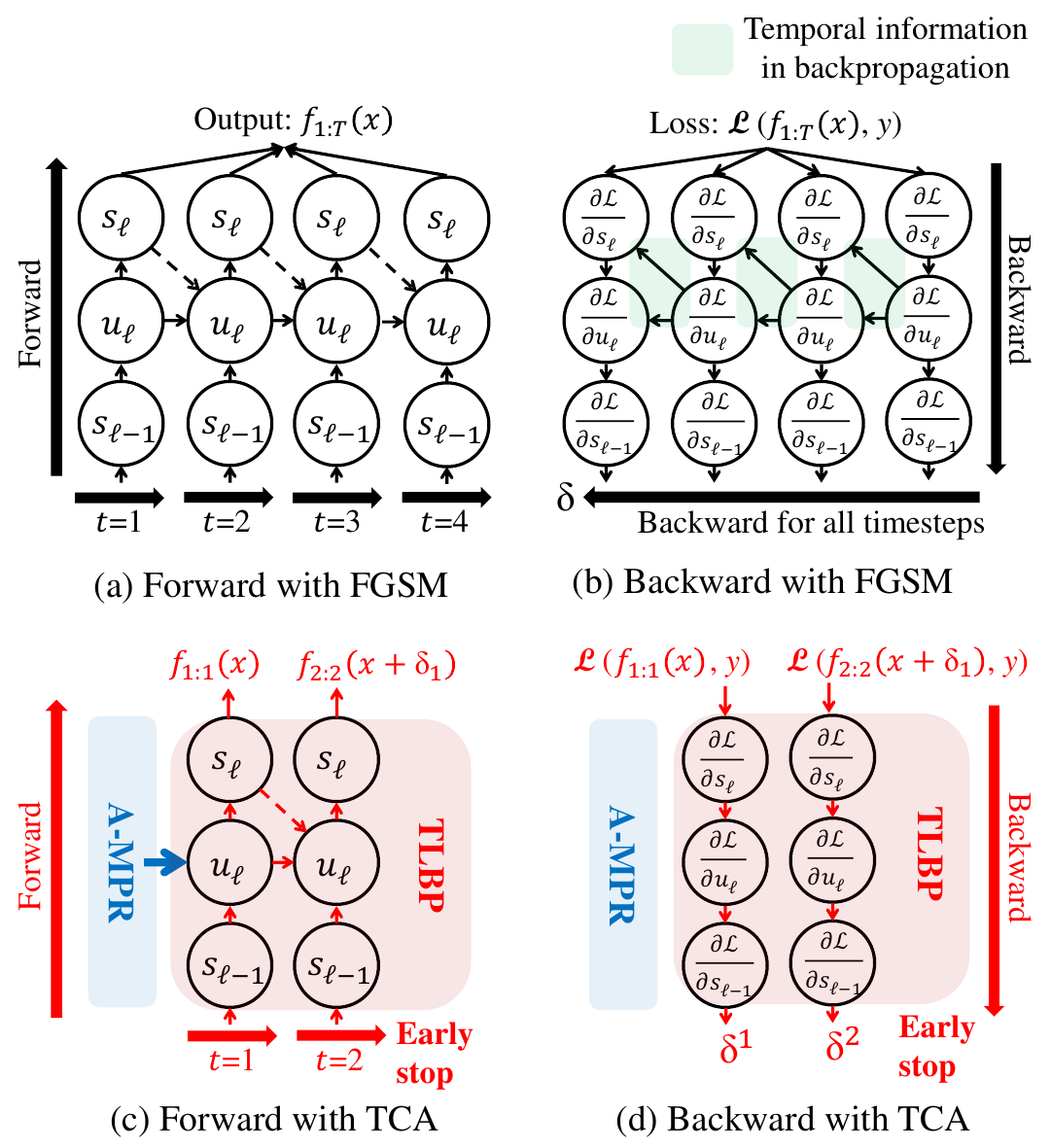}
    \caption{
    Forward pass and backward pass for FGSM (a--b) and TCA (c--d) for window size $w=1$. Since TCA does not rely on temporal information between consecutive windows, it can decide whether to stop the attack at each window using TLBP and measured cross-entropy. The blue line indicates the reuse of pre-extracted membrane potentials for each class through A-MPR for rapid warm-up.
    }
    \label{fig:tca_method}
\end{figure}

In this section, we present the foundational mechanisms of SNNs and outline the principles of adversarial attacks.
As illustrated in Fig.~\ref{fig:tca_method}(a), an SNN propagates an input image (denoted by $x$) across multiple timesteps and layers. We formulate this forward pass, denoted by $f_{1:T}$, as follows:

\begin{equation}
f_{1:T}(x) = \frac{1}{T} \sum^{T}_{t=1} h \circ l^{L} \circ l^{L-1} \circ \cdots \circ l^{1}(x),
\label{eq:func_snn}
\end{equation}
where $T$ and $L$ are the number of timesteps and blocks, respectively. 
$l^{\ell}(\cdot)$ represents the $\ell$-th block, and each block consists of a convolution layer, a batch normalization layer, and a leaky integrated-and-fire (LIF) layer~\cite{GKN14}.
$h(\cdot)$ denotes the classifier in the last layer of the network.

At each timestep, LIF neurons accumulate membrane potential across $T$.
The membrane potential $u_{\ell,t}$ at the $t$-th timestep in $\ell$-th block is given by:

\begin{equation}
    u_{\ell,t}=\tau u_{\ell,t-1}+W_{\ell}s_{\ell-1,t} + \eta_{\ell, t},
    \label{eq:membrane}
\end{equation}
where $\tau$ and $W_{\ell}$ denote the leaky factor and the convolutional weight in the $\ell$-th block, respectively~\cite{BRC07}.
$\eta_{\ell, t}$ denotes the reset factor that is applied to the membrane potential $u_{\ell,t}$ when it exceeds a certain threshold.
$s_{\ell-1,t}$ represents the spike fired at the $t$-th timestep in the $(\ell-1)$-th block.
We formulate $s_{\ell-1,t}$ as follows:

\noindent
\begin{align}
s_{\ell-1,t} &= 
\begin{cases}
1 & \text{if }u_{\ell-1,t} > V_{th}, \\
0 & \text{otherwise}
\end{cases}, %\tag{3}
\label{eq:spike}
\end{align}
where $V_{th}$ is predefined threshold.
If $u_{\ell-1,t}$ exceeds the threshold $V_{th}$, the LIF neuron fires a spike to the next layer and resets its membrane potential.

Adversarial attacks are approaches used to deceive SNNs into making incorrect predictions by generating imperceptible perturbations.
The formula to generate perturbations (illustrated in Fig.~\ref{fig:tca_method}(b)) is defined as the following optimization problem:

\begin{equation}
    \delta = \argmax_{\delta,||\delta||_{\infty} \leq \epsilon}\mathcal{L}(f_{1:T}(x+\delta|\theta), y),
    \label{eq:adver_attack}
\end{equation}
where $\delta$ is a generated perturbation and $\mathcal{L}$ is a loss function. 
$\epsilon$ is a constant value that ensures the perturbation remains imperceptible.
$f_{1:T}(\cdot|\theta)$ is an SNN model parameterized by $\theta$, and $y$ is the target label intended to mislead.

We consider two basic gradient-based attack methods applied in both scenarios: FGSM~\cite{GSS14} and PGD~\cite{KGB18}.
FGSM is an attack method that generates a perturbation by calculating the sign of the gradients against the loss function, that is

\begin{equation}
    \delta = \epsilon \cdot \text{sgn}(\nabla_{\delta}\mathcal{L}(f_{1:T}(x|\theta), y)),
    \label{eq:fgsm}
\end{equation}
where $\text{sgn}(\cdot)$ represents the function to extract sign of the gradients.
PGD, an advanced extension of FGSM, generates refined perturbations through iterative gradient extraction~\cite{MMS17}.
PGD adjusts the size of perturbations $\gamma$ for the given total iterations, which is expressed as follows:

\begin{equation}
    \delta^{r} = \text{clip}_{[-\epsilon, \epsilon]}(\delta^{r-1} + \gamma \cdot \text{sgn}(\nabla_{\delta^{r-1}}\mathcal{L}(f_{1:T}(x+\delta^{r-1}|\theta), y))), 
    \label{eq:pgd}
\end{equation}
where $r$ is the index of iteration (for $1 \le r$, $\delta^{0}=0$), and $\text{clip}(\cdot)$ ensures that the perturbation satisfies $||\delta||_{\infty} \leq \epsilon$.
However, applying these attack techniques to SNNs requires at least $T$ forward and backward propagations.

\section{\sys{}: Methodology}
\label{sec:method}

This section details \sys{}, a method targeting classification problems in both white-box and black-box settings, motivated by real-time applications (e.g., traffic sign recognition in autonomous vehicles). 
Our threat model is grounded in practical feasibility. 
In a scenario without any system access, the attack can be physically realized by projecting optical perturbations onto objects (e.g., traffic signs), a technique demonstrated by AttackZone~\cite{MMC22}. 
Alternatively, with limited physical access such as USB attachment, a man-in-the-middle (MITM) attack—where the image stream is intercepted and reinjected after perturbation—becomes feasible, as shown in recent studies~\cite{WRW24, LHH22}.

\subsection{Observations}
\label{subsec:motivation}

We present two measurement-based observations that reveal the inefficiencies of conventional SNN attacks and form the foundation for our proposed method, \sys{}.

\begin{figure}[t!]
    \centering
    \includegraphics[width=1\linewidth]{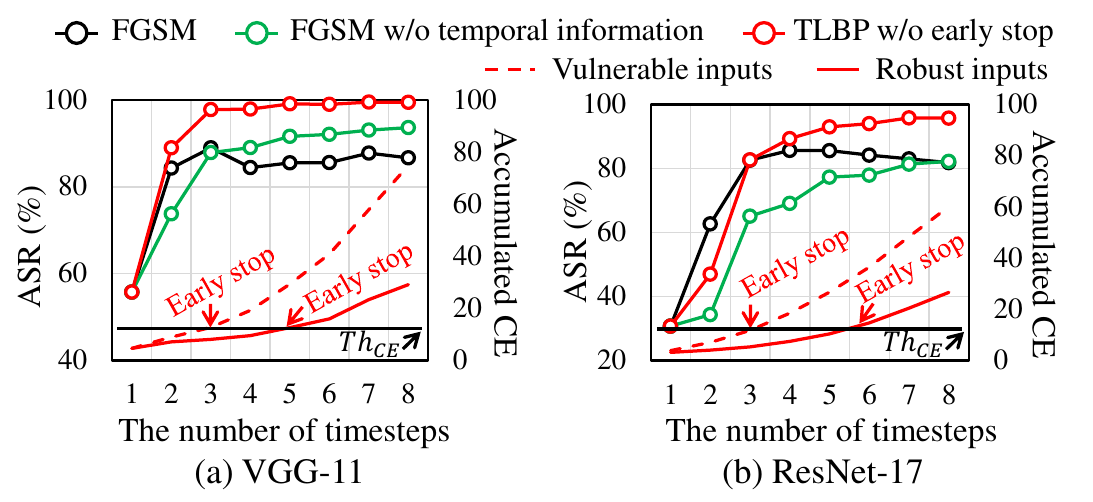}
    \caption{
    ASR and accumulated CE (calculated by $\mathcal{L}(f_{t_{s}:t_{e}}(\cdot), y)$) over timesteps on CIFAR-10 for three STBP variants (FGSM, FGSM w/o temporal information, and TLBP (w/o early stop, $w=1$)), evaluated on (a) VGG-11 and (b) ResNet-17.     
    Vulnerable and robust inputs refer to samples successfully attacked by TLBP at the first and eighth timesteps, respectively.}
    \label{fig:Timestep_ASR}
\end{figure}

\paragraph{Necessity of temporal information.}
Our first analysis questions the reliance on full-timestep processing in conventional SNN attacks. 
As shown by the standard STBP-based FGSM attack in Fig.~\ref{fig:Timestep_ASR} (black line), the ASR quickly saturates, suggesting that processing all timesteps is often redundant. 
This standard approach requires a full forward pass followed by an extensive backward pass because the membrane potential at each timestep $t$ is dependent on the state at $t-1$ (Eq.~\eqref{eq:membrane}), analogous to BPTT in RNNs\footnote{Note that we consider the direct training method for SNNs, which adapts the BPTT algorithm by using a surrogate gradient to handle the non-differentiable LIF activation function.}.

To investigate the necessity of this temporal dependency, we conducted a motivational experiment with an attack variant that disregards it (i.e., eliminating BPTT). 
This approach draws inspiration from an emerging theme in SOTA research, where shortening the temporal backpropagation path is used to mitigate challenges in direct SNN training, such as vanishing gradients. Our variant calculates gradients on a per-timestep basis, effectively treating each timestep as independent. 
Despite sacrificing this global temporal information, the resulting attack (green line in Fig.~\ref{fig:Timestep_ASR}) achieves a comparable final ASR to the standard method. This finding leads to our first key observation:

\begin{itemize}[leftmargin=6.5mm]
    \item[O1.] The global temporal information aggregated across all timesteps has a minimal impact on the final ASR of gradient-based attacks.
\end{itemize}

\begin{figure}[t!]
    \centering
    \includegraphics[width=1\linewidth]{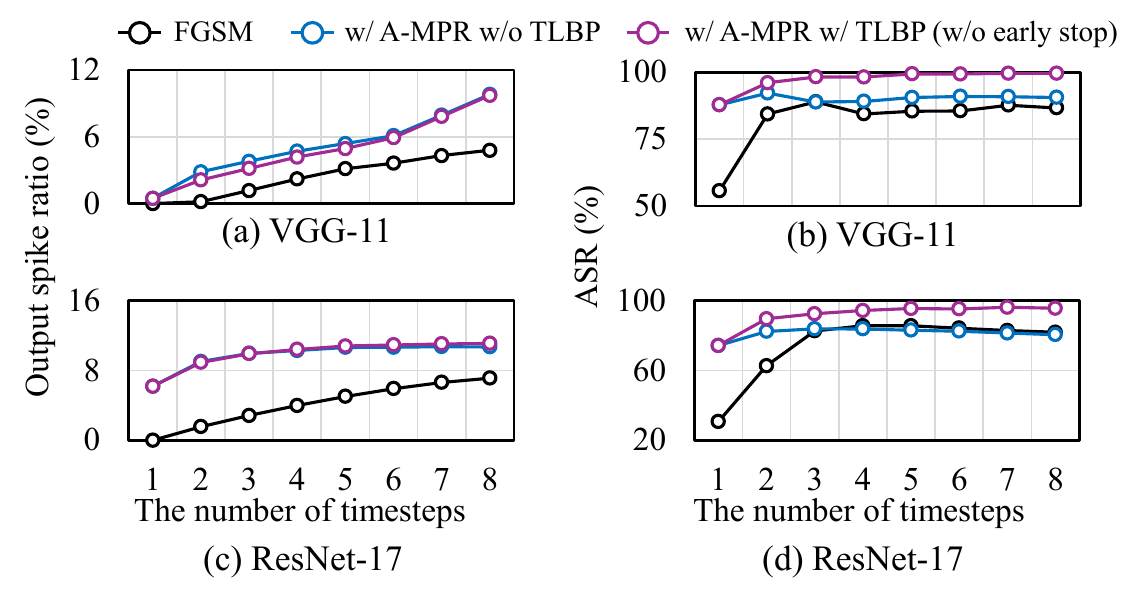}
    \caption{Output spike ratio (a, c) and ASR (b, d) for three STBP-based methods—a standard FGSM, an attack with A-MPR, and one combining A-MPR with TLBP (without early stopping, $w=1$)—evaluated over different timesteps on VGG-11 and ResNet-17 with CIFAR-10.}
    \label{fig:snn_propagation}
\end{figure}

\paragraph{Inefficacy of early timesteps.}
The second challenge arises from the initial timesteps of an attack. Fig.~\ref{fig:snn_propagation} shows that both the neuron firing rate and the corresponding ASR of FGSM (black line) are negligible at the beginning of the attack sequence. 
This indicates that insufficient information propagates through the network in the early stages, resulting in ineffective gradients. 
This forms our second observation: 

\begin{itemize}[leftmargin=6.5mm]
    \item[O2.] Gradients extracted from the initial timesteps are ineffective due to low neuron firing activity, hindering the attack process from the start.
\end{itemize}

\subsection{TLBP: Timestep-Level Backpropagation}
\label{subsec:TLBP}

Building upon O1, TLBP enables input-dependent early termination by monitoring an attack's success at runtime, a capability absent in conventional FGSM and PGD. 
Instead of processing all timesteps at once, TLBP operates on a ``window," a unit of $w$ timesteps ($1 \le w \le T$) over which a single forward-backward pass is performed to generate a perturbation. 
This windowed approach generalizes the conventional attacks; if the window size $w$ is set to the total number of timesteps $T$, a single iteration of TLBP is equivalent to FGSM, while multiple iterations are equivalent to PGD.

The operational flow of TLBP is illustrated in Fig.~\ref{fig:tca_method}(c) and (d) for a window size of $w=1$. 
In $n$-th window, the forward pass (Fig.~\ref{fig:tca_method}(c)) computes the SNN output over the current window, $f_{t_{s}:t_{e}}(x+\delta^{n-1})$. 
This output is used to calculate the CE loss, $\mathcal{L}(f_{t_{s}:t_{e}}(\cdot), y)$, which serves as the input to the backward pass (Fig.~\ref{fig:tca_method}(d)). 
The backward pass then computes the gradient to update the perturbation from $\delta^{n-1}$ to $\delta^n$ as follows:
\begin{equation}
    \delta^{n} = \text{clip}_{[-\epsilon, \epsilon]}(\delta^{n-1} + \epsilon \cdot \text{sgn}(\nabla_{\delta^{n-1}}\mathcal{L}(f_{t_{s}:t_{e}}(x+\delta^{n-1}|\theta), y))),
    \label{eq:tlbp}
\end{equation}
where $t_{s}=(n-1)\cdot w+1$ and $t_{e}=\min(n\cdot w, T)$ define the window boundaries for the $n$-th window ($1\le n$, $\delta^0=0$). 
This per-window strategy not only provides a robust criterion for early stopping but also counter-intuitively improves the final ASR compared to full-timestep methods, as evidenced by the significant performance gap between TLBP (red line with circle markers) and standard FGSM (black line) in Fig.~\ref{fig:Timestep_ASR}.

To determine when to stop, TLBP uses cross-entropy (CE) as a proxy for attack success. A high CE loss signifies that the model's confidence in the true class is low, indicating that the perturbation is successfully steering the prediction toward the adversarial target. 
The attack terminates once the accumulated CE, defined as $\sum_{n=1}\mathcal{L}(f_{t_{s}:t_{e}}(\cdot), y))$, reaches a predefined threshold $Th_{CE}$ (or when $t_{s} \ge T$ holds). 
The validity of this metric is empirically demonstrated in Fig.~\ref{fig:Timestep_ASR}, where inputs vulnerable to attacks exhibit a correspondingly rapid accumulation of CE.

\subsection{A-MPR: Adversarial Membrane Potential Reuse}
\label{subsec:A-MPR}

A-MPR is a rapid warm-up technique that addresses the inefficiency of early attack timesteps based on O2. It overcomes the ``cold start" problem of SNNs by initializing the network with pre-calculated, non-zero membrane potentials, ensuring effective gradient extraction from the onset of the attack.
The insight for A-MPR comes from observing the negligible initial spike ratios in conventional attacks (Fig.~\ref{fig:snn_propagation}(a, c)). 
This indicates a wasteful warm-up phase where neurons only accumulate potential without firing, yielding less useful gradients. 
A-MPR bypasses this inefficient runtime phase by pre-calculating and reusing these initial membrane potentials.

Since randomly generating these potentials can degrade ASR, we pre-generate ``membrane image" $y^*$ for each attack target $y$ using a purpose-built loss function that balances two objectives: (i) ensuring the generated potentials are similar to those from real inputs for gradient accuracy, and (ii) steering the initial inference away from the correct label for adversarial efficacy. 
We propose $\mathcal{L}_{mem}$ for objective (i) and $\mathcal{L}_{adv}$ for (ii). 
The similarity loss, $\mathcal{L}_{mem}$, is formulated using cosine similarity ($\text{sim}(\cdot, \cdot)$) as follows:
\begin{equation}
    \mathcal{L}_{mem}(t_{1}) = \sum^{L}_{\ell=1} \text{sim}(u_{\ell,t_{1}}, u^{*}_{\ell,t_{1}}),
    \label{eq:mem_loss}
\end{equation}
where $u_{\ell,t_{1}}$ and $u^{*}_{\ell,t_{1}}$ are the membrane potentials at timestep $t_{1}$ and block $\ell$ accumulated by the input image $x$ and the trainable membrane image $y^*$, respectively. The adversarial loss, $\mathcal{L}_{adv}$, encourages the pre-calculated potentials to already be adversarial:
\begin{equation}
    \mathcal{L}_{adv}(x, y, t_{1}) = \mathcal{L}(f_{t_{1}+1:T}(x), y).
    \label{eq:adv_loss}
\end{equation}
The total loss for generating the membrane image is the sum of these two components:
\begin{equation}
    \mathcal{L}_{mpr}(x, y, t_{1}) = \mathcal{L}_{mem}(t_{1}) + \mathcal{L}_{adv}(x, y, t_{1}).
    \label{eq:total_loss}
\end{equation}
The membrane image is then optimized via gradient ascent for each class in the training set:
\begin{equation}
    y^{*} = y^{*} + \beta \cdot \text{sgn}(\nabla_{y^{*}}\mathcal{L}(x, y, t_{1})),
    \label{eq:update_mem}
\end{equation}
where $\beta$ is the update parameter.
Before an attack, the network is warmed up by reusing the prepared $y^*$ for the target label $y$.
As a post-processing step, we clamp all membrane potentials to a range of $[V_{min}, V_{max}]$ to mitigate the effect of outliers. 
Since the entire A-MPR process is performed offline, it introduces no computational overhead to the runtime attack.

The results in Fig.~\ref{fig:snn_propagation}(b, d) validate this approach. 
A-MPR alone (blue line) significantly boosts the initial ASR over the standard FGSM (black line). 
When combined with TLBP (purple line), the ASR saturates even more rapidly, creating an ideal condition for the early stopping criterion from Section~\ref{subsec:TLBP} to effectively reduce latency. 
The comprehensive latency-ASR performance of the complete \sys{} framework is demonstrated in the evaluation section.

\section{Evaluation}
\label{sec:evaluation}

In this section, we evaluate the effectiveness of \sys{} against existing attack methods in both white-box and black-box scenarios. 
Our experiments are conducted on VGG-11~\cite{SKZ14} and ResNet-17~\cite{ZWD21} architectures with the CIFAR-10/100~\cite{KRH09} and CIFAR10-DVS~\cite{SKZ14} datasets\footnote{Results for CIFAR10-DVS are provided in the supplement due to page limits.}. 
All experiments are conducted on NVIDIA RTX A6000 GPUs.
All victim SNNs are trained using the STBP method, as ANN-SNN converted models have known security vulnerabilities. 
We compare the performance of our proposed \sys{} framework against the baseline FGSM and PGD attacks, evaluating each in combination with three SOTA gradient generation methods: STBP, RGA, and HART. 
Performance is measured using two metrics: ASR, the proportion of samples that cause a misclassification, and the attack latency, the total time required to generate a perturbation for a single input. 
\begin{figure*}[t!]
    \centering
    \includegraphics[width=0.92\linewidth]{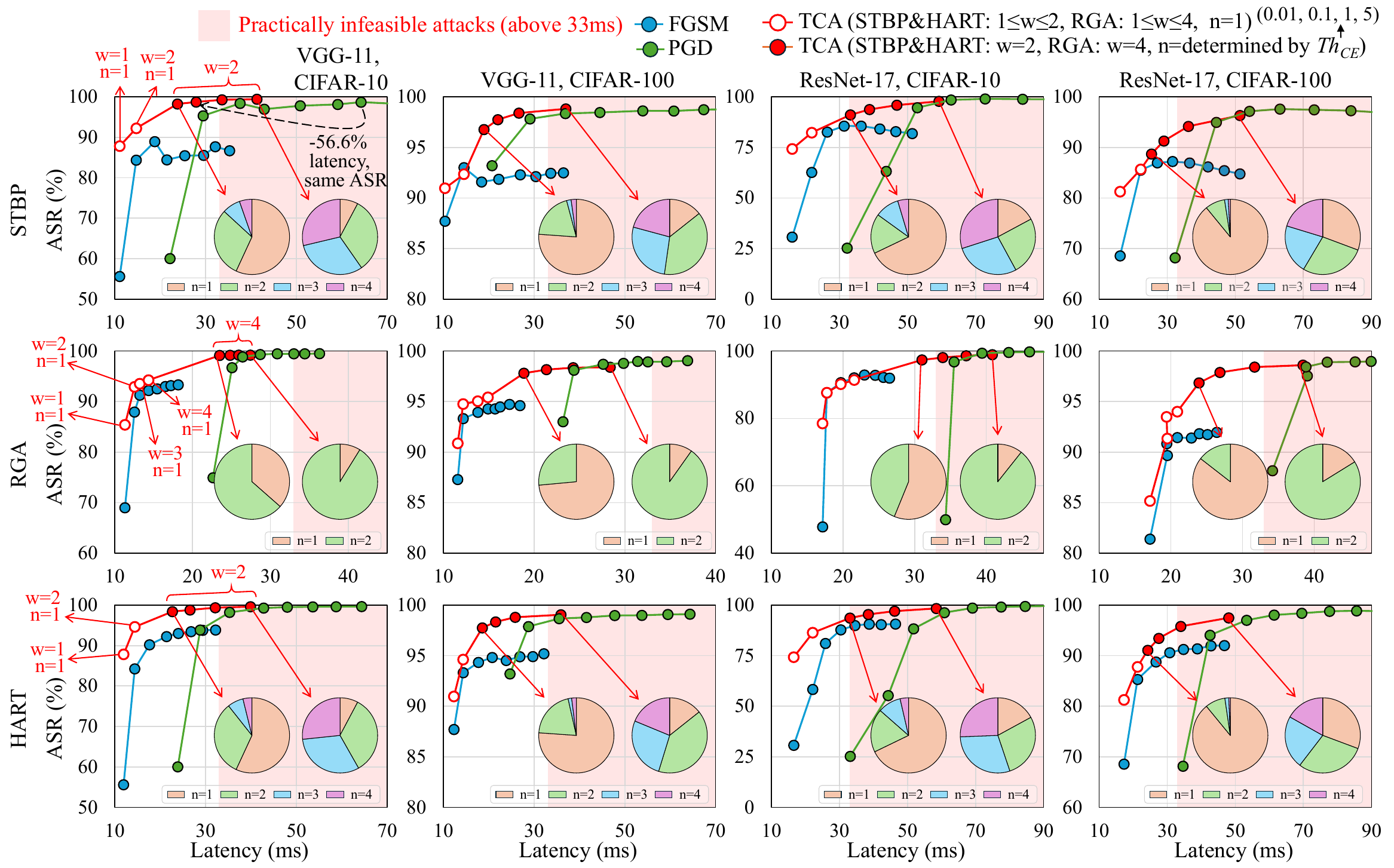}
    \caption{Experimental results on whitebox scenario}
    \label{fig:eval_white}
\end{figure*}
\begin{figure*}[t!]
    \centering
    \includegraphics[width=0.92\linewidth]{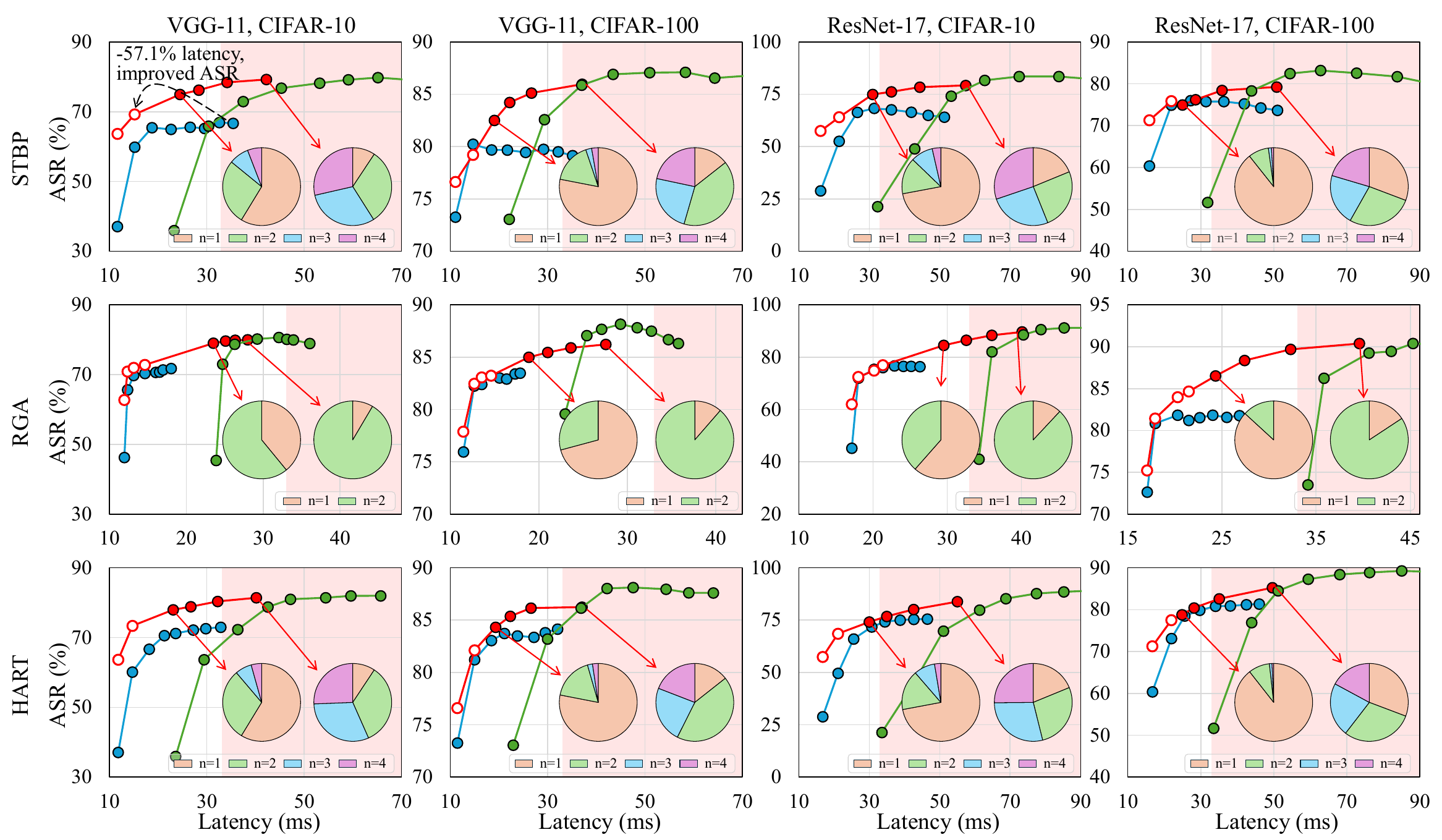}
    \caption{Experimental results on blackbox scenario}
    \label{fig:eval_black}
\end{figure*}

\subsection{Comparison to Prior Art}
\label{subsec:eval_main}
We evaluate the latency-ASR effectiveness of \sys{} against FGSM and PGD baselines in Figs.~\ref{fig:eval_white} and~\ref{fig:eval_black}. 
The experiments are conducted on VGG-11 and ResNet-17 with CIFAR-10/100 datasets in an untargeted scenario, using $\epsilon=8/255$, $t_{1}=2$, $\beta=8/255$, $V_{min}=0.2$, and $V_{max}=0.5$. 
To ensure a fair comparison at low latencies, PGD is restricted to two iterations with a step size of $8/255$. 
The points for the baselines represent attacks with a fixed number of timesteps from one to eight. 
In contrast, \sys{} operates flexibly; we plot two execution options. 
The hollow red points represent attacks using only the first window ($n=1$) with varying window sizes ($w$). 
The solid red points represent attacks with a fixed $w$ and varying $Th_{CE}$ values (0.01, 0.1, 1, or 5), which determines the number of iterations (windows) $n$. 
Based on our motivation in Sec.~\ref{sec:intro}, we consider attacks that complete within 33ms (equivalent to 30 FPS) to be practically feasible for real-time systems.

\paragraph{White-box scenario.}
As shown in Fig.~\ref{fig:eval_white}, most PGD configurations are practically infeasible (in the red square area of each subfigure) due to their iterative nature, while FGSM, though feasible, achieves a relatively low ASR. 
For \sys{}, the results (solid red points) show that $Th_{CE}$ acts as a control-knob for the latency-ASR trade-off; a lower threshold allows the attack to stop at an earlier window (e.g., $n=1$), while a higher threshold requires more windows (e.g., $n=4$). 
\sys{} demonstrates a superior latency-ASR trade-off compared to both FGSM and PGD across most settings (e.g., up to 56.6\% latency reduction for STBP on VGG-11 and CIFAR-10), while its performance on the VGG-11 and CIFAR-100 combination remains comparable.

\paragraph{Black-box scenario.}
In the black-box scenario, we assume the attacker knows only the model architecture, not any specific parameters (following the same black-box assumptions of RGA and HART for a fair comparison). 
Using the same as the white-box setting (e.g., $\epsilon$, $t_{1}$, $\cdots$, $w$, $n$), TCA again demonstrates a superior latency-ASR trade-off compared to the baselines across most configurations (e.g., up to 57.1\% latency reduction for STBP on VGG-11 and CIFAR-10), as shown in Fig.~\ref{fig:eval_black}, despite a marginally lower ASR on the VGG-11 and CIFAR-100 combination.

\subsection{Ablation Study}

\paragraph{Impact of loss function in A-MPR.}
We ablate the components of our proposed A-MPR---the adversarial loss ($\mathcal{L}_{adv}$), the similarity loss ($\mathcal{L}_{mem}$), and the class ($y$)-specific strategy---to analyze their individual impact on \sys{}'s performance. 
The results are shown in Table~\ref{tab:loss_cifar10}, which presents the ASR for VGG-11 and ResNet-17 on CIFAR-10 under STBP, RGA, and HART attacks. 
To isolate the effect of A-MPR, we fix the number of iterations to one ($n=1$) and test for window sizes $w=1$ and $w=2$. 
The results indicate that incrementally adding each component generally increases ASR, regardless of the model or backpropagation method. For instance, for VGG-11 with STBP at $w=1$, the ASR increases from 84.7\% to 87.8\% with all components enabled.

\begin{table}[t!]
\centering
\caption{Ablation study of A-MPR on CIFAR-10}
\label{tab:loss_cifar10}
\footnotesize
\resizebox{\columnwidth}{!}{
\begin{tabular}{c|ccc|cc|cc}
\hline
\multirow{2}{*}{BP} &
\multirow{2}{*}{$\mathcal{L}_{adv}$} & \multirow{2}{*}{$\mathcal{L}_{mem}$} & \multirow{2}{*}{$y$-specific} & \multicolumn{2}{c|}{VGG-11}   & \multicolumn{2}{c}{ResNet-17} \\ \cline{5-8} 
&            &            &           & $w$=$1$       & $w$=$2$       & $w$=$1$        & $w$=$2$    \\ \hline
\multirow{3}{*}{STBP}
& \ding{51}  & \ding{55}  & \ding{55} & 84.7          & 89.1          & 74.2           & 80.4          \\
& \ding{51}  & \ding{51}  & \ding{55} & 85.6          & 88.0          & \underline{\textbf{74.8}}  & 80.6          \\
& \ding{51}  & \ding{51}  & \ding{51} & \underline{\textbf{87.8}} & \underline{\textbf{92.2}} & 74.3           & \underline{\textbf{82.3}} \\ \hline \hline
\multirow{3}{*}{RGA}
& \ding{51}  & \ding{55}  & \ding{55} & 82.1          & 91.4          & 76.9           & 85.6          \\
& \ding{51}  & \ding{51}  & \ding{55} & 84.4          & 92.2          & 78.2           & 85.1          \\
& \ding{51}  & \ding{51}  & \ding{51} & \underline{\textbf{85.4}} & \underline{\textbf{92.9}} & \underline{\textbf{78.5}}  & \underline{\textbf{87.6}} \\ \hline \hline
\multirow{3}{*}{HART}
& \ding{51}  & \ding{55}  & \ding{55} & 84.0          & 92.5          & 73.0           & 84.0          \\
& \ding{51}  & \ding{51}  & \ding{55} & 85.2          & 92.4          & \underline{\textbf{74.5}}  & 85.0          \\
& \ding{51}  & \ding{51}  & \ding{51} & \underline{\textbf{87.8}} & \underline{\textbf{94.6}} & 74.3  & \underline{\textbf{86.3}} \\ \hline
\end{tabular}
}

\end{table}
\begin{table}[t!]
\centering
\caption{Various minimum values of membrane potentials on CIFAR-10}
\label{tab:mem_ablation}
\footnotesize
\begin{tabular}{c|c|cc|cc}
\hline
\multirow{2}{*}{BP} & \multirow{2}{*}{($V_{min}$, $V_{max}$)} & \multicolumn{2}{c|}{VGG-11} & \multicolumn{2}{c}{ResNet-17} \\ \cline{3-6}
&                     & $w$=$1$       & $w$=$2$       & $w$=$1$        & $w$=$2$    \\ \hline
\multirow{5}{*}{STBP}
& (none, none)        & 14.8          & 17.0          & 25.4           & 29.8          \\
& (0.0, 1.0)          & 60.8          & 82.7          & 36.8           & 55.8          \\
& (0.1, 0.8)          & 78.8          & 91.8          & 53.6           & 72.0          \\
& (0.2, 0.6)          & \underline{\textbf{89.6}}          & \underline{\textbf{93.5}}          & 71.7           & 80.6          \\
& (0.2, 0.5)          & 87.8          & 92.2          & \underline{\textbf{74.3}}           & \underline{\textbf{82.3}}          \\ \hline \hline
\multirow{5}{*}{RGA}
& (none, none)        & 29.1          & 37.9          & 36.8           & 47.5         \\
& (0.0, 1.0)          & 71.3          & 91.2          & 51.4           & 77.4         \\
& (0.1, 0.8)          & 79.0          & 93.7          & 65.1           & 85.2         \\
& (0.2, 0.6)          & \underline{\textbf{87.4}}          & \underline{\textbf{93.9}}          & 76.4           & \underline{\textbf{88.1}}         \\
& (0.2, 0.5)          & 85.4          & 92.9          & \underline{\textbf{78.5}}           & 87.6           \\ \hline \hline
\multirow{5}{*}{HART}
& (none, none)        & 14.8          & 19.0          & 25.4           & 29.6         \\
& (0.0, 1.0)          & 60.8          & 84.6          & 36.6           & 56.5         \\
& (0.1, 0.8)          & 78.8          & 92.8          & 53.2           & 75.8         \\
& (0.2, 0.6)          & \underline{\textbf{89.6}}          & \underline{\textbf{95.7}}          & 71.6           & 85.1         \\
& (0.2, 0.5)          & 87.8          & 94.6          & \underline{\textbf{74.3}}           & \underline{\textbf{86.3}}         \\ \hline
\end{tabular}
\end{table}

\paragraph{Boundary of membrane potentials.}
We investigate the impact of clamping the membrane potentials to a range of $[V_{min}, V_{max}]$ during the A-MPR process. As shown in Table~\ref{tab:mem_ablation}, not bounding the potentials~(none, none) causes a significant drop in ASR across all configurations, which we attribute to incorrectly accumulated potentials in outlier neurons. 
In contrast, setting the boundaries (e.g., $V_{min}=0.2$, $V_{max}=0.5$) substantially improves performance. 
For example, for ResNet-17 with STBP ($w=1$, $n=1$), the ASR increases from a very low 25.4\% to as high as 74.3\% when the potentials are appropriately bounded.

\begin{figure}[t!]
    \centering
    \includegraphics[width=0.9\linewidth]{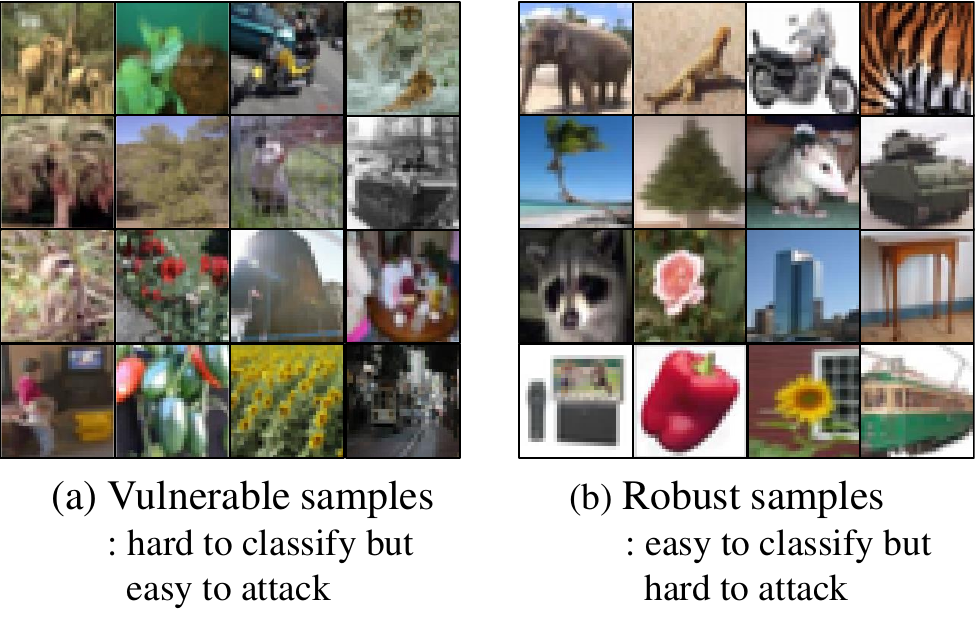}
    \caption{Visualization of vulnerable (attacked at $n=1$) and robust (attacked at $n=8$) samples on VGG-11 with CIFAR-100}
    \label{fig:input_vulnerable}
\end{figure}

\paragraph{Attack vulnerability.}
Fig.~\ref{fig:input_vulnerable} visualizes the characteristics of inputs from CIFAR-100 that are either vulnerable or robust to our attack on the VGG-11 model. 
We define `vulnerable' samples as those successfully attacked in the first window ($w=1, n=1$), and `robust' samples as those that resist the initial attack but are compromised in the $8$-th window ($w=1, n=8$). 
The visualization indicates that vulnerable samples are often visually ambiguous (e.g., small objects or poor contrast), corresponding to low initial confidence scores for the ground truth. 
Consequently, these samples require only a small perturbation over a few timesteps to be misclassified, whereas robust samples, with their high-confidence visual clarity, demand a larger, cumulative perturbation.

\section{Conclusion}
\label{sec:conclusion}

In this paper, we addressed the critical limitation of high latency in SOTA adversarial attacks on SNNs, which renders them practically infeasible for real-time systems. 
We introduced TCA, a framework built on two key insights into SNN attack dynamics. 
We first analyzed the global temporal information in the backpropagation of the SNN attack and revealed its minimal impact on an attack's final success. 
Based on this finding, we developed TLBP to perform per-window evaluation, enabling effective early stopping without sacrificing performance. 
To accelerate the initial attack phase, A-MPR bypasses the inefficient warm-up period by reusing pre-calculated potentials. 
Our experiments on VGG-11 and ResNet-17 with CIFAR-10/100 and CIFAR10-DVS validate this approach, demonstrating a reduction in required attack latency of up to 56.6\% and 57.1\% for white-box and black-box attacks, respectively, all while maintaining a comparable ASR to SOTA methods.

\bibliography{aaai2026}

\end{document}